\documentclass{article}
\usepackage{spconf,amsmath,graphicx,hyperref}
\usepackage{amssymb}
\usepackage{multirow}
\usepackage{booktabs}
\usepackage[normalem]{ulem}
\useunder{\uline}{\ul}{}

\title{Breaking Data Efficiency Dilemma: A Federated and Augmented Learning Framework For Alzheimer's Disease Detection via Speech}

\name{%
\begin{tabular}{@{}c@{}}
Xiao Wei$^{1,2}$, Bin Wen$^{1,2}$, Yuqin Lin$^{2,3}$, Kai Li$^{2}$, Mingyang Gu$^{1,2}$, \\
Xiaobao Wang$^{1}$, Longbiao Wang$^{1,4,*}$, Jianwu Dang$^{2,*}$\thanks{* Corresponding authors.}
\end{tabular}}

\address{
$^1$Tianjin Key Laboratory of Cognitive Computing and Application, \\College of Intelligence and Computing, Tianjin University, Tianjin, China\\
$^2$Shenzhen Institutes of Advanced Technology, Chinese Academy of Sciences, Shenzhen, China\\
$^3$College of Computer and Data Science, Fuzhou University, Fuzhou, China\\
$^4$Huiyan Technology (Tianjin) Co., Ltd, Tianjin, China}

\begin{document}
\ninept
\maketitle
\begin{abstract}
Early diagnosis of Alzheimer's Disease (AD) is crucial for delaying its progression. While AI-based speech detection is non-invasive and cost-effective, it faces a critical data efficiency dilemma due to medical data scarcity and privacy barriers. Therefore, we propose FAL-AD, a novel framework that synergistically integrates federated learning with data augmentation to systematically optimize data efficiency. Our approach delivers three key breakthroughs: First, absolute efficiency improvement through voice conversion-based augmentation, which generates diverse pathological speech samples via cross-category voice-content recombination. Second, collaborative efficiency breakthrough via an adaptive federated learning paradigm, maximizing cross-institutional benefits under privacy constraints. Finally, representational efficiency optimization by an attentive cross-modal fusion model, which achieves fine-grained word-level alignment and acoustic-textual interaction. Evaluated on ADReSSo, FAL-AD achieves a state-of-the-art multi-modal accuracy of 91.52\%, outperforming all centralized baselines and demonstrating a practical solution to the data efficiency dilemma. Our source code is publicly available at \url{https://github.com/smileix/fal-ad}.
\end{abstract}
\section{Introduction}

Alzheimer's Disease (AD) is a prevalent neurodegenerative disease that severely impacts patients' memory, cognition and behavior, posing a substantial burden on global public health systems. Due to its irreversible nature and prolonged preclinical stage, early diagnosis is critical for delaying disease progression and improving patients' quality of life \cite{de2020artificial}. In recent years, owing to the non-invasiveness, low cost and scalability of spontaneous speech analysis, it has emerged as a promising tool for AD screening \cite{yang2022deep}, where deep learning methods can capture subtle signs of cognitive decline, demonstrating significant potential for practical application.

High-performance deep learning diagnostic models relies heavily on large-scale, high-quality annotated speech data, however, the lack of such data is a fundamental bottleneck in current research. This bottleneck reflects a data efficiency dilemma in medical AI, the inherent data scarcity and privacy-related barriers. This challenge manifests at two interrelated levels: first, the absolute lack of efficiency, e.g., data scarcity, collecting AD speech data requires rigorous clinical coordination, resulting in high costs and long cycles, which inherently limits the amount of available data. This seriously constrains the model's learning capacity, leading to overfitting and weak generalization \cite{luz2021alzheimer}. Second, the relative efficiency loss, e.g., data silos, due to medical privacy regulations and institutional barriers, already scarce data are further fragmented across hospitals and research centers, preventing collaborative integration and scale benefits \cite{teo2024federated}. The two issues reinforce each other, creating a vicious cycle.

Existing studies have yet to systematically address this dilemma. Most mainstream efforts focus on designing more complex models (e.g., \cite{chen2021automatic, braun2024infusing}) to extract more discriminative features, or exploring multi-modal fusion to capture cross-modal interactions (e.g., \cite{cai2023exploring, chatzianastasis2023neural}, ), yet largely overlook the fundamental issue of data efficiency dilemma. Although recent work such as \cite{meerza2022fair, hsu2024cluster, ouyang2023design} has begun to explore federated learning (FL) to mitigate privacy barriers, these approaches only establish a basic collaborative framework without resolving the inherent problem of data scarcity. Moreover, most methods adopt simple late fusion strategies, failing to fully exploit the representational capacity of the data. Another study \cite{kalabakov2024comparative} also highlights that standard FL may under-perform in highly heterogeneous and scarce data environments. None of these methods offers a comprehensive solution to enhance data utilization efficacy.

\begin{figure*}[htp]
        \label{model}	
        \begin{minipage}[h]{0.33\linewidth}
			\centering
            \label{model:a}
			\centerline{\includegraphics[width=1\textwidth]{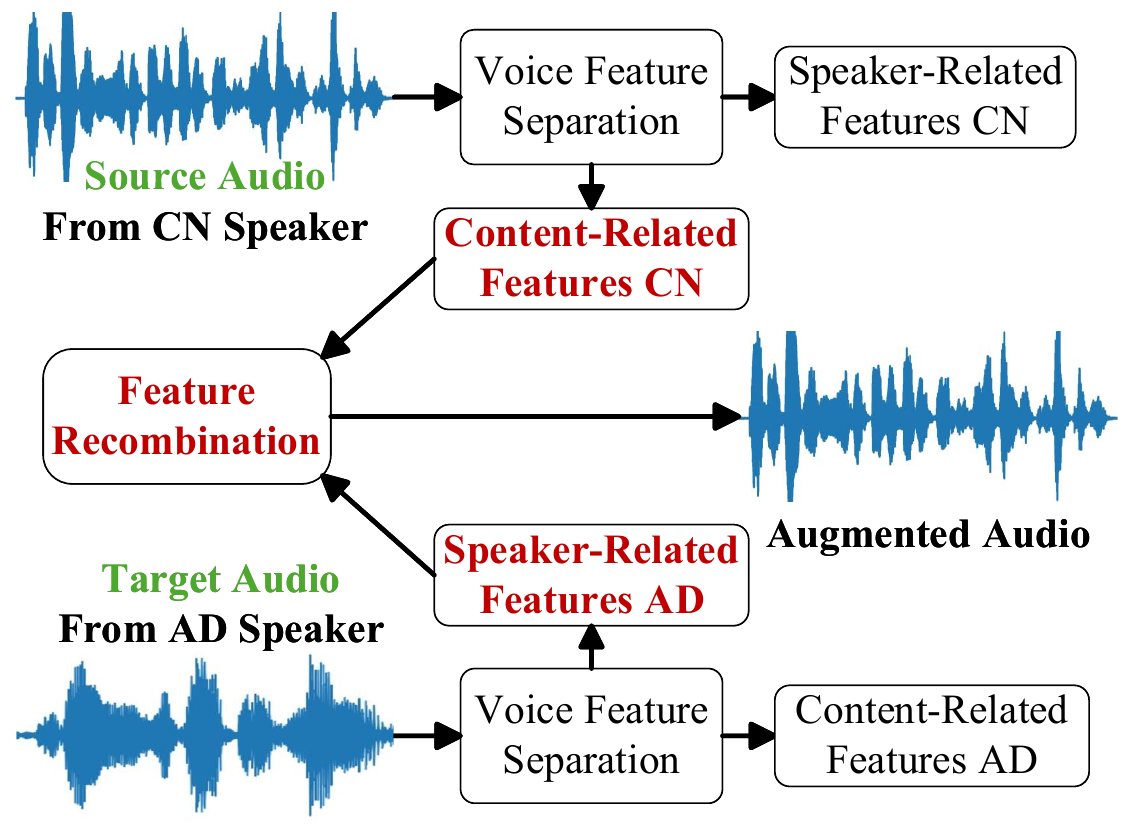}}
			\centerline{(a) Data Augmentation}\medskip
		\end{minipage}
		\hfill
		\begin{minipage}[h]{0.28\linewidth}
			\centering
            \label{model:b}
			\centerline{\includegraphics[width=1\textwidth]{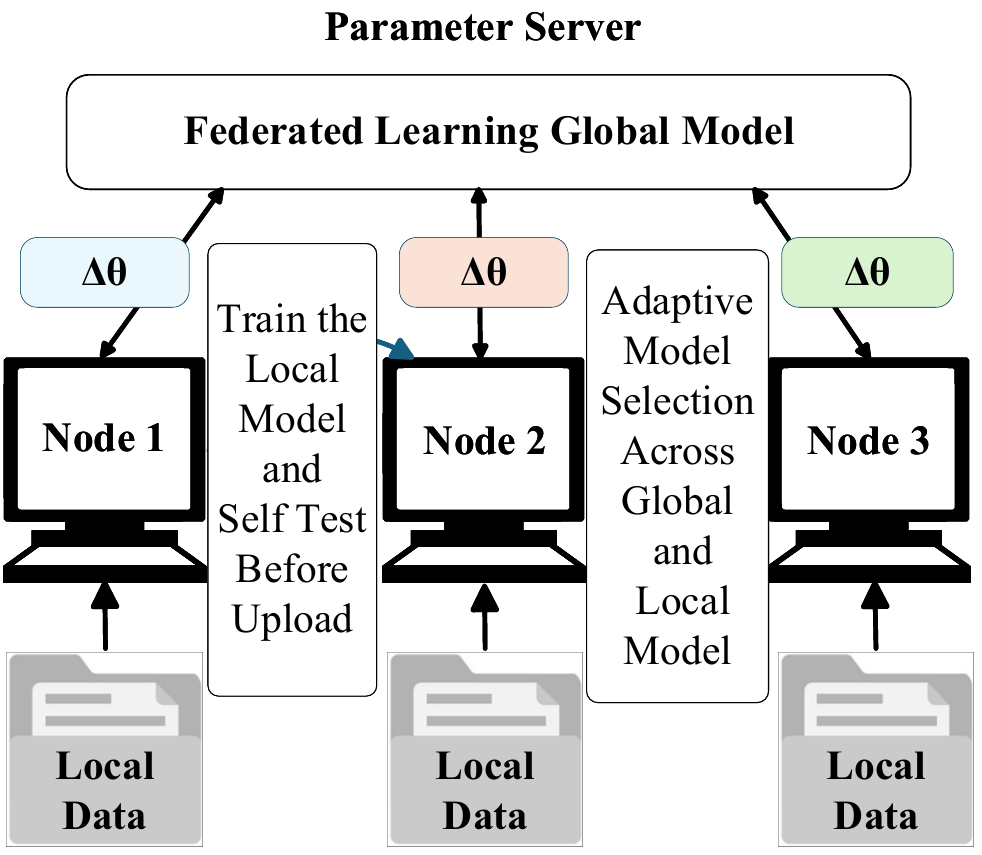}}
			\centerline{(b) Federated Learning}\medskip
		\end{minipage}
        \hfill
        \begin{minipage}[h]{0.33\linewidth}
			\centering
            \label{model:c}
			\centerline{\includegraphics[width=1\textwidth]{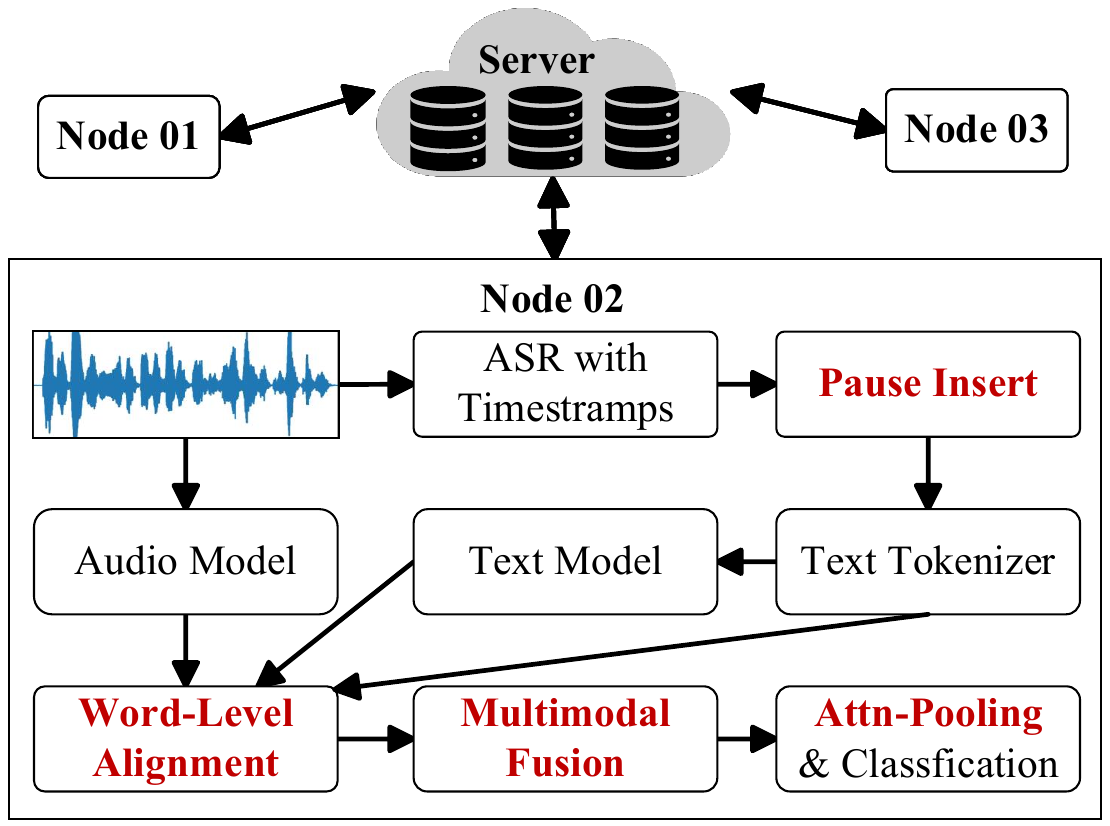}}
			\centerline{(c) Cross-Modal Fusion Model}\medskip
		\end{minipage}
     \vspace{-11pt}   
		\caption{Overview of the proposed FAL-AD framework. (a) Cross-category Voice Conversion-based Data Augmentation: generates pathological speech samples by recombining speaker features from one category and content features from another. (b) Adaptive Federated Learning: clients collaboratively train under privacy constraints and select their optimal model from the federation history. (c) Cross-Modal Fusion Model: depicts the word-level alignment and attentive fusion process of acoustic and textual features.}
		
     \vspace{-11pt}
\end{figure*}

To fundamentally address the data efficiency bottleneck, we propose the \textbf{F}ederated and \textbf{A}ugmented \textbf{L}earning framework for \textbf{A}lzheimer's \textbf{D}isease Detection  (\textbf{FAL-AD}), whose core idea is to comprehensively optimize the data efficiency from generation to utilization. To our knowledge, this is the first work to systematically explore data efficiency. Our contributions are threefold:
\begin{itemize}
    \item Absolute efficiency improvement: we construct a cross-category voice conversion-based data augmentation strategy to generate high-quality pathological samples, effectively expanding the data volume and diversity.
    \item Collaborative efficiency breakthrough: we design an adaptive federated learning paradigm that establishes a cross-silo collaboration mechanism, connecting isolated data silos into an efficient cooperative network.
    \item Representational efficiency optimization: we improve an attention-based cross-modal fusion model to achieve fine-grained alignment and deep interaction, ensuring maximal knowledge acquiring from limited data.
\end{itemize}

\section{The Federated and Augmented Framework}

\subsection{Overview}

Fig.\ref{model} illustrates the overall framework of our proposed FAL-AD, which consists of three core modules: (1) a data augmentation module, responsible for expanding the training data; (2) a federated learning module, used for collaborative model training; and (3) a cross-modal fusion module, used for final classification decision. Each module is described in detail below.

\subsection{Data Augmentation with Cross-category Voice Conversion}

To address the scarcity of AD speech data, we employ a voice data augmentation strategy. While text-to-speech (TTS) synthesis is commonly used for augmentation, TTS systems are typically trained on healthy speech samples and offer limited controllability. Even after adjusting parameters such as speech rate and pause duration, the synthesized speech inevitably exhibits artificial characteristics and fails to capture the authentic pathological patterns of real AD patients. To more effectively generate disease-relevant vocal manifestations while minimizing interference from non-pathological factors (e.g., speaker identity), we introduce voice conversion (VC) technology, as shown in Fig.\ref{model:a}(a). This method preserves the linguistic and pathological content of speech while transforming speaker-dependent characteristics, thereby enabling the model to focus on clinically relevant features.

Specifically, Our voice conversion is implemented using the CosyVoice 2.0B model \cite{du2024cosyvoice}. Let $V_i = (S_i, C_i, Y_i)$ denote an original speech sample, where $Y_i \in \{ \text{AD}, \text{CN} \}$ (CN stands for cognitively normal), $S_i$ denotes the speaker characteristics (including timbre and identity-related features), and $C_i$ represents the content aspects (including linguistic content, rhythm, pauses, and pathology-related acoustic features). During augmentation, 
for source sample $V_{i}$ and target sample $V_{j}$ with the opposite category ($Y_{i} \neq Y_{j}$), we employ voice conversion to recombine the speaker characteristics of the target audio $S_{j}$ with the content features of the source audio $C_{i}$, thereby generating a new augmented sample:

\begin{equation}
\widetilde{V}_{i,j} = \mathcal{VC}(V_i, V_j) = (S_j, C_i, Y_i),
\end{equation}
where $\mathcal{VC}(\cdot)$ denotes the voice conversion function, $Y_i \neq Y_j$, and the label of the augmented sample satisfies $\widetilde{Y}_{i,j} = Y_i$. Each sample in the dataset is alternately treated as the target sample, while a sample from the opposite category is randomly selected as the source sample to perform the aforementioned voice conversion.

Through cross-category voice conversion, we generate both positive and negative class samples for each speaker, significantly enhancing the dataset's diversity. This strategy also resolves class imbalance issues while ensuring models no longer favor majority classes. Crucially, by maintaining identical speaker timbre across both sample types, the technique forces models to concentrate exclusively on learning pathology-related acoustic features, such as AD-specific prosodic abnormalities and pause patterns, and reduces the risk of model overfitting to irrelevant speaker-specific features. This innovative decoupling of timbre and pathological traits not only ensures precise control over diagnostic characteristics but also substantially improves the model's accuracy, robustness, and generalization capability to unseen samples.

\subsection{Federated Learning with Adaptive Model Selection}

We adopt Federated Learning (FL) as the foundational paradigm for collaborative training, as Fig.\ref{model:b}(b) shows. In the standard Federated Averaging (FedAvg) procedure \cite{mcmahan2017communication}, the server executes the following steps at each communication round $r$:

\begin{enumerate}
    \item \textbf{Server Distribution}: The central server distributes the current global model parameters $\omega^r$ to all participating clients.
    \item \textbf{Local Client Update}: Each client $i$ performs local training for $E$ epochs using its local dataset $\mathcal{D}_i$, resulting in a set of local model updates $\Delta \omega_i^r$.
    \item \textbf{Model Aggregation}: The server aggregates the received model updates from the clients: 
    \begin{equation}
        \omega^{r+1} \leftarrow \omega^r + \sum\nolimits_{i=1}^{M} \frac{n_i}{n} \Delta \omega_i^r,
    \end{equation}
    where $n = \sum_{i=1}^{M} n_i$ denotes the total number of training samples across all clients, and $n_i$ is the number of samples on client $i$.
\end{enumerate}

However, the standard FL framework solely outputs the final aggregated global model $\omega^G$, potentially overlooking superior intermediate models obtained during training. To mitigate this limitation, we introduce an \textbf{adaptive model selection strategy}, which operates as follows:

\begin{enumerate}
    \item \textbf{Performance Tracking}: In each federated round $r$, after completing local training and before uploading the model to the server, each client $i$ evaluates the performance of its local model $\omega_i^r$ on a local validation set and retains a copy of the model. Concurrently, after the server completes global aggregation and distributes the new global model $\omega^{r+1}$, each client downloads this global model and evaluates the performance of the global model on the same local validation set, also saving the result.
    \item \textbf{Optimal Model Selection}: This process repeats for a predefined number of communication rounds (e.g., $R = 30$ rounds). Upon conclusion of the federated training, each client $i$ selects the model with the best performance on its local validation set from all saved model snapshots as its ultimate deployment model, denoted $\omega_i^*$.
\end{enumerate}

Our adaptive strategy maximizes performance and adaptability by dynamically selecting the optimal model $\omega_i^*$ for each client, whether a personalized local update or a generalized global model. The approach remains simple and efficient, introducing no extra hyper-parameters and adding only minimal evaluation and storage overhead. Each client thus deploys a tailored model $\omega_i^*$ that ensures high performance and local relevance.

\subsection{Attentive Cross-Modal Fusion Model}

To effectively integrate information from speech and text modalities, we build upon the CogniAlign \cite{ortiz2025cognialign} framework, as shown in Fig.\ref{model:c}(c), where we conduct targeted improvements to adapt it for the federated learning environment. The final fusion model achieves fine-grained multimodal interaction through the following mechanisms:

First, as for feature extraction and alignment, we begin with prosodic information modeling, where the Whisper \cite{radford2023robust} automatic speech recognition (ASR) model generates transcriptions with word-level timestamps. According to the timestamps, three types of pause markers (comma, period, and ellipsis) are inserted into the text sequence to capture pause and disfluency patterns \cite{yuan2020disfluencies}. Subsequently, we use pre-trained models to extract token-level textual features and frame-level acoustic features, and then according to the timestamps, the latter is mean-pooled into word level to complete alignment.

Second, as for the feature fusion, we employ a gated cross-attention mechanism, implemented through a single-layer Transformer encoder to enable deep interaction between modalities \cite{vaswani2017attention}. Specifically, the audio embeddings serve as the Query while the text embeddings provide the Key and Value. The gated cross-modal attention is computed as:
\begin{equation}
\mathbf{H}_{\text{att}} = \text{Attention}(\mathbf{A}, \mathbf{T}, \mathbf{T}), \quad \mathbf{G} = \sigma(\mathbf{W}_g \mathbf{H}_{\text{att}} + \mathbf{b}_g),
\end{equation}
\begin{equation}
\mathbf{H} = \mathbf{G} \odot \mathbf{H}_{\text{att}} + (1 - \mathbf{G}) \odot \mathbf{A},
\end{equation}
where \( \textbf{A} \) and \( \textbf{T} \) respectively represent the word-level embedding sequences of audio and text, \( \sigma \) denotes the sigmoid function, $\textbf{W}_g$ and $\textbf{B}_g$ are trainable parameters, and \( \odot \) represents element-wise multiplication. And then an attention pooling layer dynamically learns the importance of each token through the learned weights:
\begin{equation}
\alpha_i = \mathbf{W}_a \mathbf{H}_i + \mathbf{b}_a , \quad w_i = \text{softmax}(\alpha)_i,
\end{equation}
\begin{equation}
\mathbf{h} = \sum\nolimits_{i=1}^T w_i \mathbf{H}_i,
\end{equation}
where $\textbf{W}_a$ and $\textbf{b}_a$ are trainable parameters. The final classification is performed by a multilayer perceptron (MLP) on the aggregated representation \( \mathbf{h} \).

\section{Experiments}
\subsection{Dataset and Preprocessing}
We experiment with the ADReSSo Challenge dataset \cite{luz2021detecting}, which contains spontaneous speech recordings from 237 subjects (118 AD and 119 CN), describing the "Cookie Theft" picture. To ensure a fair comparison with existing works \cite{ortiz2025cognialign}, we adopt a five-fold cross-validation strategy. The dataset is randomly partitioned into five mutually exclusive folds, iteratively using one fold as the test set and the remaining four as the training set. The final reported performance metrics are the average results across all five folds. Note that only the training set is augmented. For both local and federated paradigms, the training set is further randomly divided into three parts to serve as the local data for three clients. 

\subsection{Experimental Setup}
To comprehensively evaluate the efficacy of the FAL-AD framework, we designed multiple comparative experiments from the perspective of learning paradigms, systematically measuring the trade-off between data collaboration and privacy preservation. Under each learning paradigm, model variants with three input modalities, i.e., audio, text, and multi-modal, so as to thoroughly validate the performance of different modalities under various paradigms.

In machine learning, Centralized Learning (CL) represents the ideal scenario of data sharing, and serves as the reference for the performance upper bound. Local Learning (LL) simulates the complete data silo scenario, where each client trains independently using only its local data without any collaboration, representing the performance lower bound. Federated Learning (FL) enables collaboration among clients while keeping data localized. This constitutes a rigorous comparative system. Comparing CL with LL quantifies the performance degradation caused by data silos; comparing LL with FL validates the fundamental benefits of federated collaboration; comparing standard FL with our proposed FAL-AD verifies the added value of our data augmentation and adaptive federated strategies.

\subsection{Implementation Details}

All audios are transcribed using the Whisper-large-v3 model \cite{radford2023robust} to obtain word-level timestamps. The audio is down-sampled to 16kHz. Audio and text features are respectively extracted using the frozen pre-trained models facebook/wav2vec2-base-960h \cite{baevski2020wav2vec} and distilbert-base-uncased \cite{sanh2019distilbert}, with the maximum sequence length set to 200. The multimodal fusion module employs with a hidden dimension of 768 and 12 attention heads. The classifier is a two-layer MLP. The hyper-parameter settings for CL and LL baselines are consistent with CogniAlign. Our FL methods are trained using the AdamW optimizer with a learning rate of 5e-5, weight decay of 0.01, and a fixed batch size of 64, which simulates 3 clients for 30 communication rounds. All hyper-parameters are selected through grid search to ensure optimal model performance.

\subsection{Results and Analysis}

\begin{table*}[ht!]
\centering
\caption{Performance comparison (Accuracy and F1-Score in \%) across different modalities and learning paradigms on the ADReSSo dataset. Results from previous centralized methods are compared against our implementations under Centralized Learning (CL), Local Learning (LL), and Federated Learning (FL) paradigms, both with and without data augmentation (Aug). The best value in each group is marked with an underline, and the global best value across all groups is marked with bold. Note that CL here is the strict reproduction version of CogniAlign using its source code.
}
\scalebox{0.97}{
\begin{tabular}{cc|cccc|cccccc}
\toprule
\multicolumn{2}{l|}{\multirow{2}{*}{\parbox{1.5cm}{Methods → \\Modals ↓}}} & \multicolumn{4}{c|}{Previous Work (Centralized Paradigm)} & \multicolumn{6}{c}{Our Work} \\ \cmidrule{3-12}
\multicolumn{2}{l|}{}                                                                               & C-Attn \cite{wang2021modular}  & Ying \cite{ying2023multimodal}  & Bang \cite{bang2024alzheimer}  & CogniAlign \cite{ortiz2025cognialign} & CL    & \multicolumn{1}{c|}{CL+Aug} & LL    & \multicolumn{1}{c|}{LL+Aug} & FL    & FL+Aug         \\ \midrule
\multicolumn{1}{c|}{\multirow{2}{*}{Audio}}                             & Acc                       & 75.30 & 71.20 & 69.01 & \underline{80.12}      & 74.55 & \multicolumn{1}{c|}{\underline{79.39}}  & \underline{68.69} & \multicolumn{1}{c|}{68.08}  & 83.84 & \textbf{\underline{85.05}} \\
\multicolumn{1}{c|}{}                                                   & F1                        & 76.00 & 73.10 & 70.39 & \underline{79.46}      & 73.39 & \multicolumn{1}{c|}{\underline{79.14}}  & 65.68 & \multicolumn{1}{c|}{\underline{67.10}}  & 83.67 & \textbf{\underline{84.64}} \\ \midrule
\multicolumn{1}{c|}{\multirow{2}{*}{Text}}                              & Acc                       & 73.50 & 78.90 & 83.10 & \underline{86.77}      & 84.85 & \multicolumn{1}{c|}{\underline{86.67}}  & 78.39 & \multicolumn{1}{c|}{\underline{79.80}}  & 87.68  & \textbf{\underline{90.30}} \\
\multicolumn{1}{c|}{}                                                   & F1                        & 73.50 & 79.00 & 83.10 & \underline{86.59}      & 84.69 & \multicolumn{1}{c|}{\underline{86.63}}  & 77.48 & \multicolumn{1}{c|}{\underline{79.55}}  & 87.64  & \textbf{\underline{90.28}} \\ \midrule
\multicolumn{1}{c|}{\multirow{2}{*}{Both}}                       & Acc                       & 77.20 & 83.70 & 87.32 & \underline{90.36}      & 86.06 & \multicolumn{1}{c|}{\underline{86.67}}  & 78.59 & \multicolumn{1}{c|}{\underline{80.61}}  & 89.70 & \textbf{\underline{91.52}} \\
\multicolumn{1}{c|}{}                                                   & F1                        & 77.60 & 83.30 & 87.25 & \underline{90.11}      & 85.89 & \multicolumn{1}{c|}{\underline{86.64}}  & 77.16 & \multicolumn{1}{c|}{\underline{80.35}}  & 89.65 & \textbf{\underline{91.45}} \\ 
\bottomrule
\end{tabular}}
\label{res1}
\vspace{-11pt}
\end{table*}

\begin{table}[ht!]
\centering
\caption{Performance comparison (Accuracy and F1-Score in \%) of federated learning algorithms with different model section strategy: standard (sFL), personalized (pFL), and adaptive (aFL) federated learning. The best performance across section strategies is bold, while the best performance across FL algorithms is underlined.}
\scalebox{0.95}{
\begin{tabular}{c|cc|cc|cc}
\toprule
\multirow{2}{*}{Algorithms} & \multicolumn{2}{c|}{sFL}  & \multicolumn{2}{c|}{pFL}  & \multicolumn{2}{c}{aFL}                     \\ \cmidrule(l){2-7} 
                        & Acc         & F1          & Acc         & F1          & Acc                  & F1                   \\ \midrule
FedAdam                 & {\ul 90.91} & {\ul 90.88} & 90.00       & 89.97       & \textbf{90.91}       & \textbf{90.88}       \\ \midrule
FedAdagrad              & 89.70       & 89.58       & 88.79       & 88.74       & \textbf{90.00}       & \textbf{89.95}       \\ \midrule
FedYogi                 & 87.27       & 87.26       & 88.48       & 88.40       & \textbf{88.79}       & \textbf{88.71}       \\ \midrule
FedProx                 & 89.70       & 89.63       & 90.60       & 90.52       & \textbf{90.91}       & \textbf{90.84}       \\ \midrule
FedAvg                  & {\ul 90.91} & 90.85       & {\ul 90.81} & {\ul 90.75} & {\ul \textbf{91.52}} & {\ul \textbf{91.45}} \\ \bottomrule
\end{tabular}}
\label{res2}
\vspace{-11pt}
\end{table}

Overall, our proposed FAL-AD framework achieves state-of-the-art performance on the ADReSSo dataset, as shown in Table~\ref{res1}, surpassing all existing centralized benchmarks. The experimental results systematically validate its effectiveness across three key aspects: first, the comparison between CL and LL quantifies the severe performance degradation caused by data isolation, highlighting the necessity of collaborative paradigms. Second, FL effectively mitigates this issue, achieving a performance gain of over 10 percentage points compared to isolated training (LL), demonstrating its capability to break down data silos while preserving privacy. Finally, and most significantly, our FL+Aug approach not only matches but exceeds the performance of its centralized counterpart (CL+Aug), achieving a new multi-modal accuracy benchmark of 91.52\% and outperforming previous state-of-the-art methods including CogniAlign (90.36\%). This paradigm breakthrough reveals a synergistic effect: federated aggregation acts as a regularizer to reduce overfitting, while data augmentation enhances generalization through expanded diversity.

\textbf{Analysis on Data Augmentation:} We further analyze the utility of the proposed voice conversion strategy under different learning paradigms. This strategy brings consistent performance improvements in almost all settings. The benefit is most prominent under the Federated Learning (FL) paradigm, boosting multimodal accuracy from 89.70\% to 91.52\%. This indicates that data augmentation provides richer and more diverse local data 'fuel' for federated learning, synergizing with the regularization effect of federated aggregation to push model performance to new heights. The gains under the Local Learning (LL) paradigm are also significant (multimodal accuracy increases from 78.59\% to 80.61\%), confirming its value as an effective regularization in extremely data-scarce environments. Notably, the improvement in CL paradigm is relatively limited (+0.61\%), consistent with our previous finding that models trained on sufficient data are more susceptible to overfitting to the subtle biases introduced by augmented data. This contrast, in turn, demonstrates that our federated learning framework is a superior environment for deploying such data augmentation strategies.

\textbf{Analysis on Federated Learning:} The performance of our FAL-AD surpasses the strong centralized baselines. This may be because that, while centralized training has access to all data, it can easily lead the model to overfit to specific patterns in the training set. The inherent "local update \& global average" iterative mechanism in the FedAvg algorithm effectively constrains the model optimization path, forcing it to converge to a flatter and more robust optimum for all data distributions, thereby achieving superior generalization performance. In addition, the adaptive model selection strategy ensures each client finally deploys its historical best model, maximizing the potential of models from different communication rounds. As shown in Table \ref{res2}, while different aggregators exhibit varying affinities for not identically and independently distributed (non-IID) data, our proposed adaptive Federated strategy (aFL) demonstrates significant and consistent advantages. It achieves optimal performance under most configurations, validating its effectiveness as a lightweight, hyperparameter-free personalization solution. Notably, traditional local fine-tuning strategies (pFL) did not bring stable gains, suggesting that mandatory local fine-tuning in heterogeneous data environments may lead the model to deviate from the optimal solution. This proves that aFL possesses strong fault tolerance and stability. The dynamic selection mechanism retains the deployment simplicity of standard FL (sFL) while achieving personalization benefits close to pFL.


\section{Conclusion}
This paper presents FAL-AD, a novel framework that tackles the data efficiency dilemma in Alzheimer's disease detection through federated learning and voice conversion-based augmentation. Our approach systematically addresses data scarcity and privacy constraints by enhancing absolute data efficiency through cross-category voice recombination, improving collaborative efficiency via adaptive federated learning with personalized model selection, and optimizing representational efficiency using attentive cross-modal fusion. Experimental results demonstrate that FAL-AD not only overcomes the limitations of data silos but achieves state-of-the-art performance with 91.52\% accuracy, surpassing existing centralized baselines while maintaining privacy preservation. In the future, we plan to explore more advanced generative augmentation techniques and sophisticated personalization methods for non-IID data scenarios, to validate the framework's generalization across multi-institutional datasets.
\newpage
\section{Acknowledgments}
This work was supported by the National High-end Talent Support Fund (No. E43301), High-end Talent Matching Fund of the Chinese Academy of Sciences (No. E55304), Guangdong Province Matching Fund for National High-end Talents (No. E47611), and National Natural Science Foundation Foundation of China (No. 62276185 and No. U23B2053).

\bibliographystyle{IEEEbib}
\bibliography{strings,refs}

\begin{thebibliography}{10}

\bibitem{de2020artificial}
Sofia De~la Fuente~Garcia, Craig~W Ritchie, and Saturnino Luz,
\newblock ``Artificial intelligence, speech, and language processing approaches to monitoring alzheimer’s disease: a systematic review,''
\newblock {\em Journal of Alzheimer’s Disease}, vol. 78, no. 4, pp. 1547--1574, 2020.

\bibitem{yang2022deep}
Qin Yang, Xin Li, Xinyun Ding, Feiyang Xu, and Zhenhua Ling,
\newblock ``Deep learning-based speech analysis for alzheimer’s disease detection: a literature review,''
\newblock {\em Alzheimer's Research \& Therapy}, vol. 14, no. 1, pp. 186, 2022.

\bibitem{luz2021alzheimer}
Saturnino Luz, Fasih Haider, Sofia de~la Fuente~Garcia, Davida Fromm, and Brian MacWhinney,
\newblock ``Alzheimer's dementia recognition through spontaneous speech,'' 2021.

\bibitem{teo2024federated}
Zhen~Ling Teo, Liyuan Jin, Nan Liu, Siqi Li, Di~Miao, Xiaoman Zhang, Wei~Yan Ng, Ting~Fang Tan, Deborah~Meixuan Lee, Kai~Jie Chua, et~al.,
\newblock ``Federated machine learning in healthcare: A systematic review on clinical applications and technical architecture,''
\newblock {\em Cell Reports Medicine}, vol. 5, no. 2, 2024.

\bibitem{chen2021automatic}
Jun Chen, Jieping Ye, Fengyi Tang, and Jiayu Zhou,
\newblock ``Automatic detection of alzheimer’s disease using spontaneous speech only,''
\newblock in {\em Interspeech}, 2021, vol. 2021, p. 3830.

\bibitem{braun2024infusing}
Franziska Braun, Sebastian~P Bayerl, Florian Hoenig, Hartmut Lehfeld, Thomas Hillemacher, Tobias Bocklet, Korbinian Riedhammer, et~al.,
\newblock ``Infusing acoustic pause context into text-based dementia assessment,''
\newblock in {\em Interspeech}, 2024, pp. 1980--1984.

\bibitem{cai2023exploring}
Hongmin Cai, Xiaoke Huang, Zhengliang Liu, Wenxiong Liao, Haixing Dai, Zihao Wu, Dajiang Zhu, Hui Ren, Quanzheng Li, Tianming Liu, et~al.,
\newblock ``Exploring multimodal approaches for alzheimer's disease detection using patient speech transcript and audio data,''
\newblock {\em arXiv preprint arXiv:2307.02514}, 2023.

\bibitem{chatzianastasis2023neural}
Michail Chatzianastasis, Loukas Ilias, Dimitris Askounis, and Michalis Vazirgiannis,
\newblock ``Neural architecture search with multimodal fusion methods for diagnosing dementia,''
\newblock in {\em ICASSP 2023-2023 IEEE International Conference on Acoustics, Speech and Signal Processing (ICASSP)}. IEEE, 2023, pp. 1--5.

\bibitem{meerza2022fair}
Syed Irfan~Ali Meerza, Zhuohang Li, Luyang Liu, Jiaxin Zhang, and Jian Liu,
\newblock ``Fair and privacy-preserving alzheimer's disease diagnosis based on spontaneous speech analysis via federated learning,''
\newblock in {\em 2022 44th Annual International Conference of the IEEE Engineering in Medicine \& Biology Society (EMBC)}. IEEE, 2022, pp. 1362--1365.

\bibitem{hsu2024cluster}
Wei-Tung Hsu, Chin-Po Chen, Yun-Shao Lin, and Chi-Chun Lee,
\newblock ``A cluster-based personalized federated learning strategy for end-to-end asr of dementia patients,''
\newblock in {\em Proc Interspeech}, 2024, vol. 2024, pp. 2450--2454.

\bibitem{ouyang2023design}
Xiaomin Ouyang,
\newblock ``Design and deployment of multi-modal federated learning systems for alzheimer's disease monitoring,''
\newblock in {\em Proceedings of the 21st Annual International Conference on Mobile Systems, Applications and Services}, 2023, pp. 612--614.

\bibitem{kalabakov2024comparative}
Stefan Kalabakov, Monica Gonzalez-Machorro, Florian Eyben, Bj{\"o}rn~W Schuller, and Bert Arnrich,
\newblock ``A comparative analysis of federated learning for speech-based cognitive decline detection,''
\newblock in {\em Proc. Interspeech 2024}, 2024, pp. 2455--2459.

\bibitem{du2024cosyvoice}
Zhihao Du, Qian Chen, Shiliang Zhang, Kai Hu, Heng Lu, Yexin Yang, Hangrui Hu, Siqi Zheng, Yue Gu, Ziyang Ma, et~al.,
\newblock ``Cosyvoice: A scalable multilingual zero-shot text-to-speech synthesizer based on supervised semantic tokens,''
\newblock {\em CoRR}, 2024.

\bibitem{mcmahan2017communication}
Brendan McMahan, Eider Moore, Daniel Ramage, Seth Hampson, and Blaise~Aguera y~Arcas,
\newblock ``Communication-efficient learning of deep networks from decentralized data,''
\newblock in {\em Artificial intelligence and statistics}. PMLR, 2017, pp. 1273--1282.

\bibitem{ortiz2025cognialign}
David Ortiz-Perez, Manuel Benavent-Lledo, Javier Rodriguez-Juan, Jose Garcia-Rodriguez, and David Tomás,
\newblock ``Cognialign: Word-level multimodal speech alignment with gated cross-attention for alzheimer’s detection,''
\newblock {\em Knowledge-Based Systems}, vol. 329, pp. 114264, 2025.

\bibitem{radford2023robust}
Alec Radford, Jong~Wook Kim, Tao Xu, Greg Brockman, Christine McLeavey, and Ilya Sutskever,
\newblock ``Robust speech recognition via large-scale weak supervision,''
\newblock in {\em International conference on machine learning}. PMLR, 2023, pp. 28492--28518.

\bibitem{yuan2020disfluencies}
Jiahong Yuan, Yuchen Bian, Xingyu Cai, Jiaji Huang, Zheng Ye, and Kenneth Church,
\newblock ``Disfluencies and fine-tuning pre-trained language models for detection of alzheimer's disease.,''
\newblock in {\em Interspeech}, 2020, vol. 2020, pp. 2162--6.

\bibitem{vaswani2017attention}
Ashish Vaswani, Noam Shazeer, Niki Parmar, Jakob Uszkoreit, Llion Jones, Aidan~N Gomez, {\L}ukasz Kaiser, and Illia Polosukhin,
\newblock ``Attention is all you need,''
\newblock {\em Advances in neural information processing systems}, vol. 30, 2017.

\bibitem{luz2021detecting}
Saturnino Luz, Fasih Haider, Sofia de~la Fuente, Davida Fromm, and Brian MacWhinney,
\newblock ``Detecting cognitive decline using speech only: The adresso challenge,''
\newblock in {\em INTERSPEECH 2021}. ISCA, 2021.

\bibitem{baevski2020wav2vec}
Alexei Baevski, Yuhao Zhou, Abdelrahman Mohamed, and Michael Auli,
\newblock ``wav2vec 2.0: A framework for self-supervised learning of speech representations,''
\newblock {\em Advances in neural information processing systems}, vol. 33, pp. 12449--12460, 2020.

\bibitem{sanh2019distilbert}
V~Sanh,
\newblock ``Distilbert, a distilled version of bert: smaller, faster, cheaper and lighter.,''
\newblock in {\em Proceedings of Thirty-third Conference on Neural Information Processing Systems (NIPS2019)}, 2019.

\bibitem{wang2021modular}
Ning Wang, Yupeng Cao, Shuai Hao, Zongru Shao, and KP~Subbalakshmi,
\newblock ``Modular multi-modal attention network for alzheimer's disease detection using patient audio and language data.,''
\newblock in {\em Interspeech}, 2021, vol. 2021, pp. 3835--3839.

\bibitem{ying2023multimodal}
Yangwei Ying, Tao Yang, and Hong Zhou,
\newblock ``Multimodal fusion for alzheimer’s disease recognition,''
\newblock {\em Applied Intelligence}, vol. 53, no. 12, pp. 16029--16040, 2023.

\bibitem{bang2024alzheimer}
Jeong-Uk Bang, Seung-Hoon Han, and Byung-Ok Kang,
\newblock ``Alzheimer's disease recognition from spontaneous speech using large language models,''
\newblock {\em ETRI Journal}, vol. 46, no. 1, pp. 96--105, 2024.

\end{thebibliography}

\end{document}